\documentclass[runningheads]{llncs}

\usepackage{eccv}

\usepackage{eccvabbrv}

\usepackage{graphicx}
\usepackage{booktabs}

\usepackage[accsupp]{axessibility}  %

\usepackage{hyperref}

\usepackage{orcidlink}

\begin{document}

\title{AlignMorph: Tuning-Free Diffusion Image Morphing via Explicit Semantic Transport}

\titlerunning{AlignMorph}

\author{Wuyi Liu\inst{1} \and
Xu Han\inst{1} \and
Yuren Chen\inst{2}\and
Yige Mao\inst{3}\and
Zishuo Peng\inst{4}\and
Xianzhi Li\inst{1}\thanks{Corresponding author}
}

\authorrunning{W.~Liu et al.}

\institute{Huazhong University of Science and Technology \and
Beijing Jiaotong University \and
Beihang University\and
Peking University\\
}

\maketitle

\begin{figure}[ht!]
    \centering
    \includegraphics[width=\linewidth]{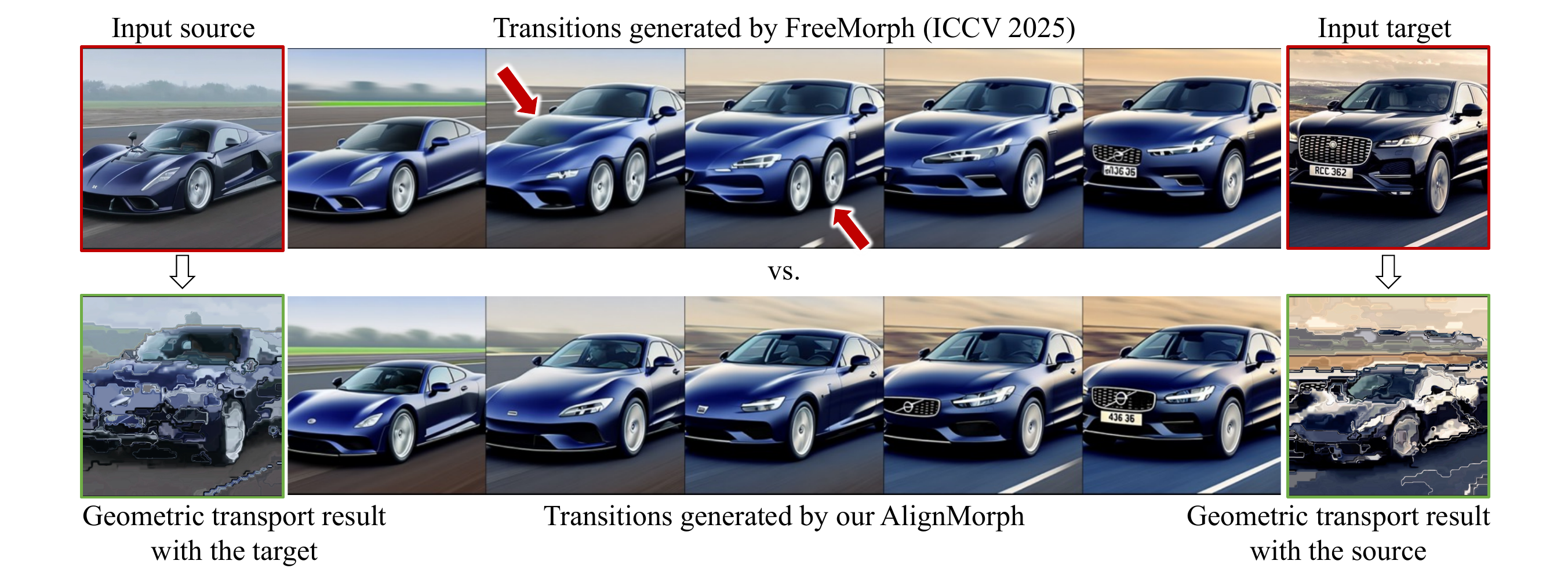}
\caption{
Comparisons between our AlignMorph and the very recent FreeMorph~\cite{cao2025freemorph} given two distinct inputs.
See the two images marked by green boxes, our explicit semantic alignment achieves accurate geometry transport at the semantic level instead of traditional pixel-level matching, leading to smoother and more reasonable transitions.
}

    \label{fig:Teaser}
\end{figure}

\begin{abstract}

Image morphing aims to produce a smooth and semantically consistent transition between two input images. Existing diffusion-based morphing methods either require expensive per-pair optimization or rely on implicit spatial alignment, which easily fails under large layout discrepancies. To address these limitations, we propose \textbf{AlignMorph}, a novel tuning-free diffusion framework guided by the principle of \emph{transport-then-denoise}. We explicitly decouple geometric alignment from generative denoising to avoid structural entanglement. Our framework consists of two core components. (1) Global Semantic Transport, which achieves diffusion-compatible semantic alignment via entropic optimal transport and reliability-aware latent warping; and (2) Coordinate-Aligned Generation, which uses a symmetric bi-phase attention handoff to maintain consistent spatial coordinates throughout denoising. Without any tuning, AlignMorph effectively eliminates ghosting and achieves superior structural coherence and temporal smoothness on morphing benchmarks. Code is available at \url{https://github.com/51xOne/Alignmorph}.

\end{abstract}

\section{Introduction}
\label{sec:intro}

Image morphing aims to generate a smooth visual transition between two distinct images through geometric warping and color blending, serving as a pivotal technique for animation and creative workflows~\cite{wolberg1998image}.
Unlike standard text-to-video generation, image morphing is jointly constrained by the input source and target images.
It requires not only smooth and continuous transitions between intermediate frames, but also semantic consistency with both input images.

To address this challenging task, generative diffusion models have been increasingly adopted as the dominant engines, building upon their extensive capabilities in high-quality image synthesis, editing, and video generation~\cite{rombach2022high,podell2023sdxl}.
Recent diffusion-based morphing methods can be broadly categorized into optimization-based and tuning-free approaches.
Optimization-based methods such as DiffMorpher~\cite{zhang2024diffmorpher} and IMPUS~\cite{yang2023impus} achieve strong fidelity by optimizing LoRA modules or latent embeddings for each image pair, but the per-instance tuning incurs substantial computational cost and limits practical use.
To remove this bottleneck, tuning-free methods represented by FreeMorph~\cite{cao2025freemorph} generate transitions by interpolating endpoint latents (e.g., via Slerp~\cite{cao2025freemorph}) and injecting endpoint information through attention. While efficient, FreeMorph implicitly assumes that the endpoints are approximately \emph{spatially aligned}.

However, this assumption does not always hold. When there exist large layout discrepancies between the input source and target images (as shown in \cref{fig:Teaser}), implicit methods cannot fully guarantee accurate spatial alignment between the two endpoint images.
Accordingly, methods such as FreeMorph may produce distorted latent trajectories during latent interpolation, which further leads to ghosting artifacts, translucent overlays, and temporal incoherence in the intermediate images generated after diffusion denoising.
Please refer to the image indicated by the arrow in \cref{fig:Teaser} as an example.

To resolve this structural entanglement, the most intuitive remedy is to explicitly align the two endpoints before interpolation and mixing. However, achieving such effective alignment is non-trivial.
Directly warping RGB images or relying on traditional low-level optical flow often fails, since the source and target typically share no pixel-level correspondence under severe layout discrepancies.
To address this issue, we argue that the explicit alignment should satisfy two critical properties.
\textbf{1): Semantic-aware alignment.}
The alignment should be conducted at a higher semantic level rather than low-level pixels, so as to ensure structural coherence between the two input images.
\textbf{2): Diffusion-compatible alignment.}
Direct spatial warping on diffusion latents would break the inherent noise statistics required for stable denoising, resulting in unstable sampling and distorted textures.
Thus, the alignment mechanism must be naturally compatible with the diffusion model.

To satisfy these dual requirements, we decouple the spatial alignment from the generative refinement. Instead of forcing the diffusion model to implicitly resolve severe layout discrepancies during sampling, we establish a reliable geometric correspondence guided by semantic features beforehand. This naturally leads to our core design principle: transport geometry first, then denoise.Guided by this principle, we propose AlignMorph, a novel tuning-free diffusion-based image morphing framework that operates in two decoupled stages.

First, \textbf{Global Semantic Transport} establishes geometric alignment between endpoints.
We estimate dense semantic correspondences via entropic optimal transport, derive reliability-gated warp fields, and apply a reliability-aware multiband latent transport that moves structural components while preserving stochastic variations.
Together, these steps align endpoint representations into a shared coordinate frame in a diffusion-compatible manner, addressing large-displacement misalignment without corrupting noise statistics.
The two images marked by green bounding boxes in \cref{fig:Teaser} visualize the geometric alignment from source to target and vice versa, showing that our method achieves semantically consistent geometry alignment rather than naive pixel-level correspondence.
Second, \textbf{Coordinate-Aligned Generation} performs diffusion sampling under explicit coordinate consistency.
We employ a symmetric bi-phase attention handoff that switches between source-aligned and target-aligned memory banks at the morph midpoint, ensuring that queries and memories remain in the same coordinate frame throughout denoising and preventing cross-frame retrieval conflicts.

Extensive experiments on large-displacement morphing benchmarks show that AlignMorph consistently improves structural integrity and temporal smoothness over existing works, while requiring no further tuning. Please see \cref{fig:Teaser} as a comparison example.
In summary, our contributions are threefold:
\begin{itemize}
    \item We propose AlignMorph, a tuning-free diffusion morphing framework based on a \emph{transport-then-denoise} paradigm, which explicitly decouples geometric alignment from generative refinement.
    \item We propose Global Semantic Transport mechanism. It achieves reliable geometric alignment by leveraging optimal-transport-based dense semantic correspondence to drive a reliability-aware multiband latent transport, aligning unaligned latents into a shared coordinate frame without corrupting diffusion noise statistics.
    \item We propose a coordinate-aligned generation mechanism with a symmetric bi-phase handoff, ensuring that queries and memory features remain in the same coordinate frame throughout denoising.
\end{itemize}

\section{Related Work}

\subsection{Image Morphing}

Image morphing aims to generate smooth visual transitions between two endpoint images. Traditional methods~\cite{beier1992feature, lee1997scattered, wolberg1998image, bookstein2002principal} typically decompose morphing into correspondence estimation and warping-based interpolation, but require manual intervention and struggle with complex appearance changes. Learning-based approaches~\cite{siarohin2019first, niemeyer2021giraffe, tian2021good} learn morphing transformations end-to-end but lack generalization to arbitrary image pairs.

Recent diffusion models~\cite{ho2020denoising, rombach2022high, podell2023sdxl, dhariwal2021diffusion} enable high-quality generative morphing. Optimization-based methods~\cite{yang2023impus, zhang2024diffmorpher} achieve strong fidelity via per-pair tuning (e.g., LoRA~\cite{hu2022lora}), but incur prohibitive computational costs. Tuning-free methods~\cite{cao2025freemorph, wang2023interpolating, kye2025chimera} eliminate this bottleneck through latent interpolation and attention feature injection, which implicitly assume spatial alignment between endpoints. Under severe spatial misalignment, this assumption fails: unaligned operations blindly entangle foreground and background semantics, causing ghosting and temporal incoherence~\cite{liu2026interp3d}. We overcome this limitation by establishing explicit semantic correspondence via optimal transport prior to diffusion sampling, effectively decoupling geometric transport from generative refinement.

\subsection{Geometric Alignment in Latent Space}

Establishing correspondences across images has been extensively studied through optical flow~\cite{horn1981determining, lucas1981iterative, sun2018pwc, teed2020raft, jiang2021learning}, feature matching~\cite{lowe2004distinctive, barnes2009patchmatch}, and semantic correspondence methods~\cite{min2019hyperpixel, truong2020glu, hong2022cost, cho2021cats, min2019spair}. Recent works leverage self-supervised representations (e.g., DINO~\cite{caron2021emerging}, DINOv2~\cite{oquab2023dinov2}) enhanced with high-resolution feature upsampling~\cite{fu2024featup} for robust matching despite appearance variations. Optimal transport (OT)~\cite{peyre2019computational} provides a principled framework for soft matching, with entropic regularization~\cite{cuturi2013sinkhorn} enabling efficient computation. Several works~\cite{liu2010sift, rocco2017convolutional, flamary2017optimal, rocco2018neighbourhood} have applied OT to vision tasks.

However, applying geometric transport to diffusion-based morphing introduces unique challenge. Diffusion latents encode both semantic structure (low-frequency) and stochastic noise (high-frequency)~\cite{choi2021ilvr, balaji2022ediff, song2020denoising}. Naively warping the entire latent corrupts high-frequency noise distributions, leading to blurred textures and unstable sampling. While frequency-domain analysis has been explored in generative models~\cite{karras2019style, karras2020analyzing, karras2021alias}, no prior work explicitly addresses the tension between geometric alignment and noise preservation in diffusion latent space. Our frequency-aware multiband latent transport resolves this by incorporating reliability-gated warping and spectral decomposition: we aggressively transport low-frequency structural components while selectively preserving high-frequency statistics, enabling geometric alignment without corrupting diffusion priors.

\subsection{Attention Mechanisms for Generative Consistency}

Attention mechanisms have become central to diffusion-based synthesis. Cross-attention enables conditional generation~\cite{rombach2022high, ramesh2022hierarchical, saharia2022photorealistic} by injecting textual or visual conditions into denoising. Recent works~\cite{zhang2023adding, mou2024t2i, brooks2023instructpix2pix} extend this with spatial control signals (e.g., ControlNet~\cite{zhang2023adding}). For image editing, methods such as Prompt-to-Prompt~\cite{hertz2022prompt}, Plug-and-Play~\cite{tumanyan2023plug}, and MasaCtrl~\cite{cao2023masactrl} manipulate attention maps for tuning-free editing. For morphing, FreeMorph~\cite{cao2025freemorph} adapts cross-attention using endpoint features as memory banks. However, under severe spatial misalignment, this induces structural entanglement: mixing unaligned memories blindly aggregates geometrically incompatible features, producing ghosting and incoherence. Video diffusion models~\cite{ho2022video, singer2022make} incorporate temporal attention but assume short-range continuity, failing under such severe misalignment. To resolve this structural entanglement, our coordinate-aligned generation mechanism employs a bi-phase handoff that explicitly switches the coordinate frame at the temporal midpoint, ensuring queries and memories share a consistent geometric space throughout denoising.

\section{Methodology}
\label{sec:method}

\begin{figure}[t]
  \centering
  \includegraphics[width=\textwidth]{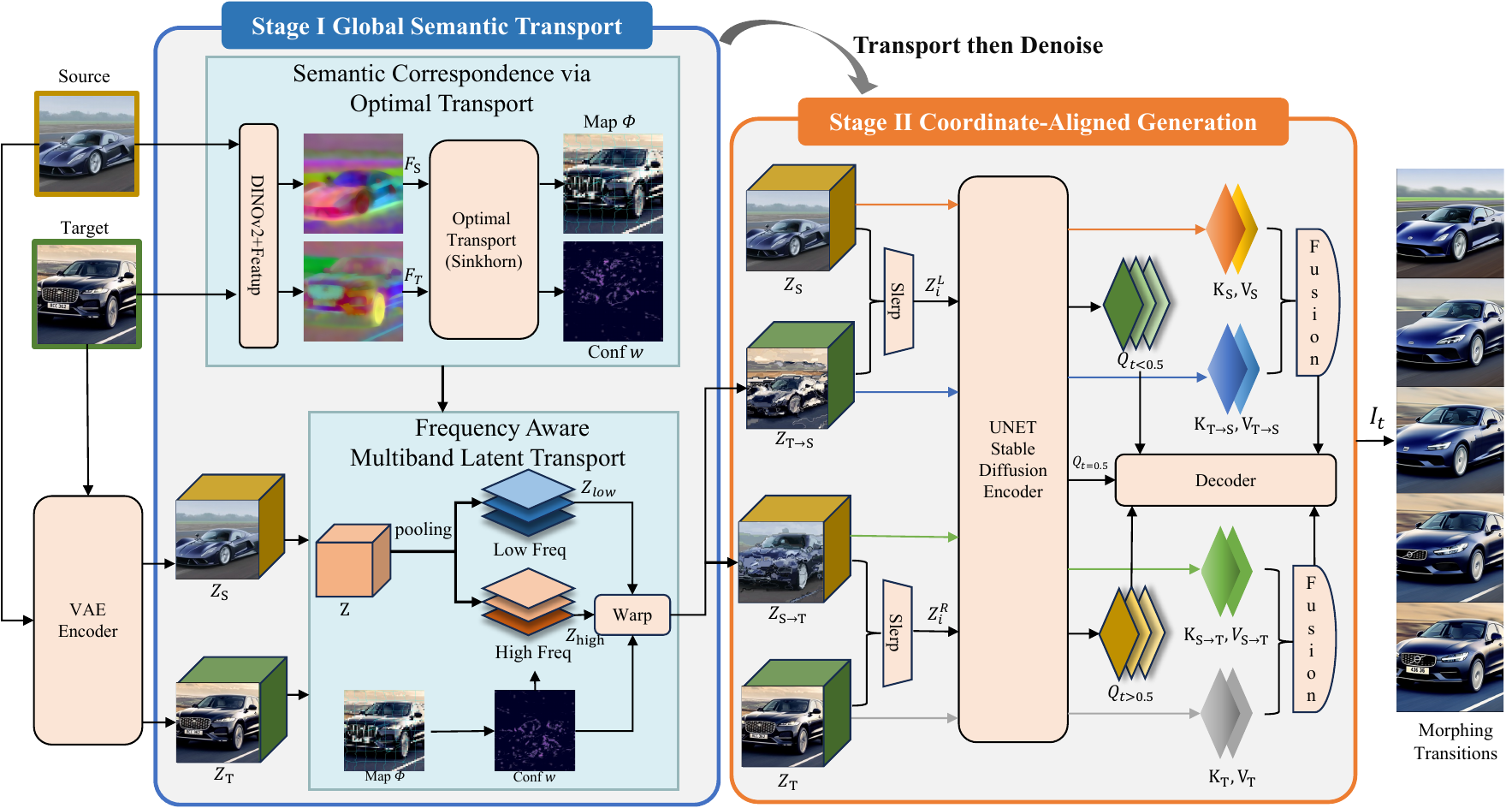}
  \caption{Overview of the AlignMorph framework. Following a \emph{transport-then-denoise} paradigm, our pipeline explicitly decouples geometric alignment from generative refinement. \textbf{Stage I} (left) establishes a shared coordinate frame: it first extracts Semantic Correspondence via Optimal Transport, and then applies Frequency-Aware Multiband Latent Transport to align endpoint latents without corrupting noise statistics. \textbf{Stage II} (right) executes Coordinate-Aligned Generation, utilizing a symmetric bi-phase attention handoff during UNet denoising to enforce geometric consistency and synthesize artifact-free transitions.}
  \label{fig:pipeline}
\end{figure}

To synthesize a smooth intermediate sequence $\{I_t\}_{t\in(0,1)}$ between two unaligned endpoints $S$ and $T$ without per-pair optimization, AlignMorph introduces a novel \emph{transport-then-denoise} paradigm that performs explicit semantic transport first by establishing semantic-level correspondence between $S$ and $T$, thereby guiding robust image morphing via diffusion denosing.
\Cref{fig:pipeline} illustrates the overall pipeline of AlignMorph, which comprises the following two stages.
(i) \textbf{Global Semantic Transport}: We first obtain semantic-aware dense correspondences via optimal transport (Sec.~\ref{sec:ot}). These correspondences are subsequently utilized to explicitly align the endpoint representations into a shared coordinate frame through a frequency-aware multiband latent transport (Sec.~\ref{sec:spectral}).
(ii) \textbf{Coordinate-Aligned Generation} (Sec.~\ref{sec:aligned_attn}): To generate structurally coherent intermediate transitions, this stage enforces geometric consistency throughout the diffusion denoising process via a symmetric bi-phase attention handoff.

\subsection{Semantic Correspondence via Optimal Transport}
\label{sec:ot}

Given two endpoint images $S$ and $T$, the primary and crucial step is to establish the correspondence between the two images, which lays the foundation for subsequent alignment and morphing.
As discussed earlier, since $S$ and $T$ often exhibit large layout discrepancies, this correspondence should be both \emph{semantic-aware} and \emph{diffusion-compatible}.
Below, we shall gradually describe how we establish such correspondences to satisfy the above two requirements.

\paragraph{(1) For semantic-aware property.}
Our goal is to build correspondence at a high semantic level rather than the pixel level.
To this end, we formulate the semantic correspondence as an Optimal Transport (OT) problem in the feature space.
Specifically, as shown in the top-left of \cref{fig:pipeline}, we first extract dense semantic features $F_S\in\mathbb{R}^{H_S\times W_S\times d}$ and $F_T\in\mathbb{R}^{H_T\times W_T\times d}$ using a ViT encoder enhanced by FeatUp~\cite{fu2024featupmodelagnosticframeworkfeatures}.
For computational efficiency, we compute the transport plan on a downsampled grid. Let $\{x_i\}_{i=1}^{N_S}$ and $\{y_j\}_{j=1}^{N_T}$ denote the resulting token sets, where $x_i,y_j\in\mathbb{R}^d$, $N_S=H_S W_S$, and $N_T=H_T W_T$. Each token is explicitly associated with its 2D spatial coordinate: $p_i\in\mathbb{R}^2$ for $x_i$ and $q_j\in\mathbb{R}^2$ for $y_j$.

To establish a structurally coherent mapping between these unaligned feature sets, we formulate the dense matching task as the OT problem. Intuitively, we view the spatial tokens of each image as two sets of points with equal weight. Matching these images is equivalent to transporting the semantic features from the source to the target. The "transportation cost" for assigning a source token $x_i$ to a target token $y_j$ is determined by their semantic dissimilarity. We explicitly measure this using the cosine distance:
\begin{equation}
C_{ij}=1-\frac{\langle x_i,y_j\rangle}{\|x_i\|_2\|y_j\|_2}
\label{eq:cost}
\end{equation}
where $\langle \cdot, \cdot \rangle$ denotes the inner product (dot product) of the two vectors, and $\|\cdot\|_2$ represents their $L_2$ norm (magnitude).

Consequently, finding a reliable dense correspondence simply translates to finding a global transport plan $T^\star\in\mathbb{R}_+^{N_S\times N_T}$ that minimizes the total transportation cost, ensuring that overall structural coherence is maintained without collapsing into many-to-one mapping errors. We solve this balanced entropic OT problem using log-domain Sinkhorn iterations to obtain $T^\star$. Finally, to convert this dense probabilistic plan into a deterministic geometric warp field, we apply barycentric projection. Specifically, by row-normalizing the transport plan as $\bar{T}_{ij}=T^\star_{ij}/(\sum_{k}T^\star_{ik}+\delta)$, we define the forward coordinate map as:
\begin{equation}
\Phi_{S\rightarrow T}(p_i)=\sum_{j}\bar{T}_{ij}\,q_j
\label{eq:bary}
\end{equation}
Here, $\Phi_{S\rightarrow T}$ is a coordinate map defined on the feature grid rather than a pixel-valued image, and the backward map $\Phi_{T\rightarrow S}$ is obtained analogously.

Note that, soft OT may be unreliable in ambiguous or occluded regions. To filter out unreliable matches, we compute a forward-backward consistency error:
\begin{equation}
\Delta(p_i)=\big\|p_i-\Phi_{T\rightarrow S}(\Phi_{S\rightarrow T}(p_i))\big\|_2.
\label{eq:fb_err}
\end{equation}
We then define a reliability weight $w(p_i)\in[0,1]$ by combining OT confidence and cycle consistency. Specifically, we use a plan sharpness score (row-max mass) to form a confidence value $c(p_i)\in[0,1]$, and apply a Gaussian penalty to the consistency error $\Delta(p_i)$:
\begin{equation}
w(p_i)=c_{S\rightarrow T}(p_i)\cdot c_{T\rightarrow S}(\Phi_{S\rightarrow T}(p_i))
\cdot \exp\!\Big(-\frac{\Delta(p_i)^2}{2\sigma^2}\Big).
\label{eq:mask}
\end{equation}

\paragraph{(2) For diffusion-compatible property.}
Having established the semantic correspondence $\Phi_{S\rightarrow T}$ and its reliability weight $w$, we must appropriately project this mapping into the latent space.
Specifically, considering that the diffusion sampling operates on a coarse latent grid of size $H_z\times W_z$, we thus project the model-space correspondence $\Phi_{S\rightarrow T}$ to this latent space by converting the coordinate map into a displacement (flow) field and resizing it. Let $r(p)=\Phi_{S\rightarrow T}(p)-p$ be the model-space displacement on the source grid. We bilinearly resize $r$ to the latent resolution and rescale it to latent pixel units to obtain $r^z(u)$, yielding the latent warp operator and reliability mask:
\begin{equation}
\Phi^z_{S\rightarrow T}(u)=u+r^z(u),
\qquad
w^z(u)=\mathrm{Resize}(w)(u).
\label{eq:latent_proj}
\end{equation}
We compute the opposite-direction latent map and reliability mask, denoted as $\Phi^z_{T\rightarrow S}$ and $w^z_{T\rightarrow S}$, analogously. The forward mask is denoted as $w^z_{S\rightarrow T}$.

\begin{figure}[t]
    \centering
    \includegraphics[width=\textwidth]{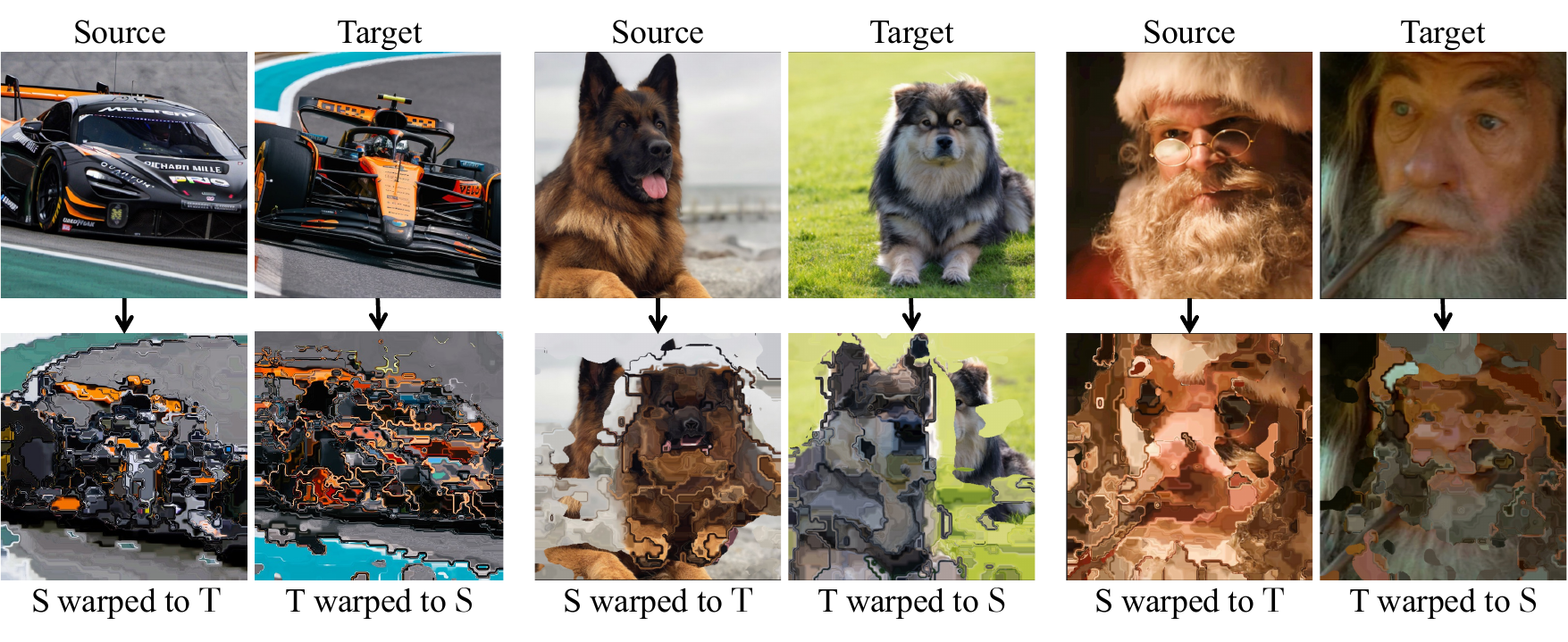}
    \caption{Bidirectional semantic warping via optimal transport. Top row: Original source and target image pairs. Bottom row: Directly decoded warped latents.}
    \label{fig:warp}
\end{figure}

To validate the effectiveness of the proposed semantic alignment, we decode and visualize the directly warped latents, as illustrated in \cref{fig:warp}. As can be observed, our approach behaves exactly as theoretically expected: it successfully extracts robust, part-level semantic correspondences rather than superficial pixel matching. Although these directly decoded latents are not perfectly realistic, they reveal a crucial capability. For instance, in the vehicle examples, the transport plan explicitly maps specific structural components such as wheels to wheels and rear wings to rear wings, despite severe pose variations and distinct environmental contexts (e.g., the green versus blue race tracks). Similarly, fine-grained features (e.g., the dog's facial parts) are routed to their correct anatomical positions. This robust, fine-grained structural alignment provides a geometrically coherent initialization, effectively resolving the spatial mismatch that fundamentally undermines traditional attention-based morphing.

\subsection{Frequency-Aware Multiband Latent Transport}
\label{sec:spectral}

Having obtained the latent warp operator $\Phi^z$ and reliability mask $w^z$, the subsequent step is to utilize them to geometrically align the endpoint diffusion latents.
The conventional approach is standard spatial warping, which treats the diffusion latent $z$ as a monolithic signal and applies a single geometric transform to all components. However, a fundamental challenge arises: this rigid treatment fails to reconcile macroscopic structural alignment with the preservation of fine-grained details and noise statistics. While low-frequency components (capturing pose and layout) can be transported reliably, high-frequency components (capturing texture details and stochastic noise) are highly sensitive to misalignment. Forcing the same warp onto these high-frequency signals disrupts their local statistics, inevitably leading to texture degradation (e.g., blurring, broken patterns) and unstable denoising.
To address this issue, we propose Frequency-Aware Multiband Latent Transport. Its core idea is to explicitly decouple geometric alignment from texture synthesis by aggressively transporting low-frequency structural components while selectively preserving high-frequency noise.

Specifically, we decompose the latent tensor $z$ into a low-frequency structural component and a high-frequency residual:
\begin{equation}
z = z_{\text{low}} + z_{\text{high}}.
\end{equation}

We obtain $z_{\text{low}}$ via spatial average pooling (kernel size $k{=}9$, stride $1$ with replicate padding), which suppresses high-frequency variations while preserving coarse semantic layout. The residual
\begin{equation}
z_{\text{high}} = z - z_{\text{low}}
\end{equation}
encodes fine-grained textures and stochastic noise variations that should not be aggressively warped in unreliable regions.

Let $\Phi^z$ denote the projected latent-grid warp field (Sec.~\ref{sec:ot}), and let $w^z\in[0,1]$ be the corresponding reliability weight on the latent grid obtained from \cref{eq:latent_proj}. We denote by $\mathcal{W}(\cdot;\Phi^z)$ a warping operator that maps each latent location according to $\Phi^z$. Our transported latent $z'$ is defined as
\begin{equation}
z'=
\underbrace{\mathcal{W}(z_{\text{low}};\Phi^z)}_{\text{Geometric guidance}}
\;+\;
\underbrace{w^z\cdot \mathcal{W}(z_{\text{high}};\Phi^z) + (1-w^z)\cdot z_{\text{high}}}_{\text{Statistical preservation}}.
\label{eq:multiband}
\end{equation}

Using \cref{eq:multiband}, we construct geometrically aligned endpoint latents:
\begin{equation}
\label{eq:transport}
z_{T\rightarrow S}=\mathcal{T}(z_T;\Phi^z_{T\rightarrow S},w^z_S),
\qquad
z_{S\rightarrow T}=\mathcal{T}(z_S;\Phi^z_{S\rightarrow T},w^z_T),
\end{equation}
where $\mathcal{T}(\cdot)$ denotes the multiband transport operator in \cref{eq:multiband}, and $w^z_S, w^z_T$ are the reliability weights derived from cycle-consistency on the corresponding coordinate frames.

\begin{figure}[t]
    \centering
    \includegraphics[width=\linewidth]{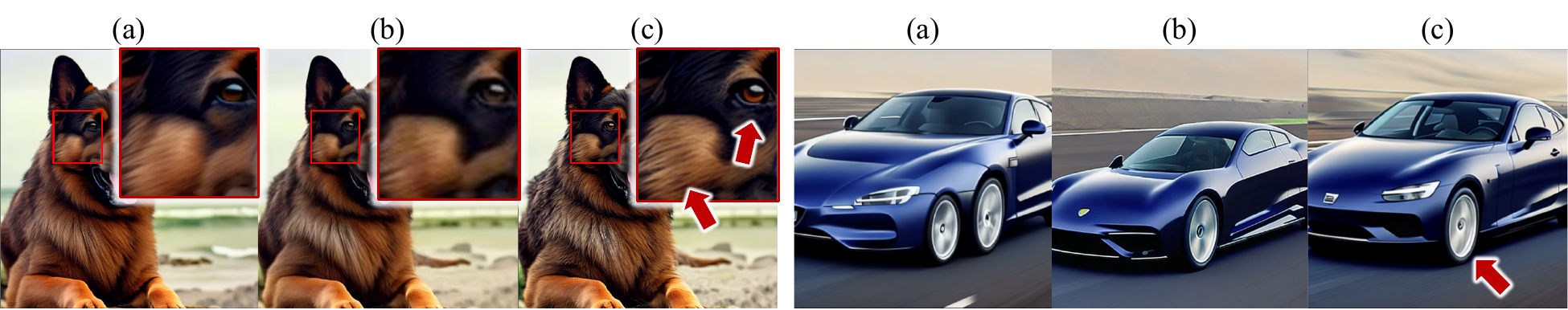}
    \caption{Comparison among (a) FreeMorph~\cite{cao2025freemorph}, (b) full latent warping (Warp-All), and (c) our multiband transport on two examples.
    }
    \label{fig:multiband}
\end{figure}

To validate the necessity of our frequency-aware multiband transport, we analyze two representative morphing scenarios in \cref{fig:multiband}: one where FreeMorph produces reasonable transitions (Left), and another where it fails completely with severe ghosting artifacts (Right). In both cases, applying full latent warping (Warp-All) indiscriminately alters the latent representation. This rigid operation disrupts the local high-frequency statistics essential for detail synthesis, causing blurred fine-grained textures (e.g., the dog's fur) and destroying specific semantic details (e.g., car headlights and sharp body edges). By explicitly decoupling frequencies, our multiband transport selectively aligns the low-frequency structural layout while strictly preserving the original high-frequency noise. Consequently, our approach eliminates ghosting where baselines fail, and maintains sharp visual fidelity that even exceeds the baseline in successful cases.

\subsection{Coordinate-Aligned Generation}
\label{sec:aligned_attn}

With the obtained geometrically aligned endpoint latents $z_{T\rightarrow S}$ and $z_{S\rightarrow T}$ via multiband transport, the next step is to synthesize the intermediate transition frames without disrupting this structural consistency during diffusion denoising.
The conventional approach in existing tuning-free methods is to interpolate unaligned endpoint latents to form the intermediate query $Q_{i,k}^l$, and directly mix unaligned endpoint memories to form the $K/V$ branches. However, this approach forces the attention mechanism to aggregate features across disparate geometric layouts, inevitably causing severe structural entanglement even before the attention is calculated.
To address this issue, we propose Coordinate-Aligned Generation. Its core idea is to explicitly construct both the interpolated query and the mixed endpoint memories within a phase-consistent coordinate frame, thereby guaranteeing that spatial indices consistently refer to the same semantic structures and making the attention fusion geometrically coherent.

Specifically, let $N$ be the number of frames and $i\in\{0,\dots,N-1\}$ the frame index.
We denote morph progress by $\alpha_i=\frac{i}{N-1}$.
Let $k\in\{1,\dots,K\}$ be the diffusion step, and
$z_S,z_T\in\mathbb{R}^{C\times H_z\times W_z}$ be endpoint latents.
Using the latent-grid transport results from \cref{eq:transport},
we build two directional latent paths via spherical linear interpolation (Slerp) from\cite{cao2025freemorph}:
\begin{equation}
z_i^{L}=\mathrm{Slerp}(z_S,z_{T\rightarrow S},\alpha_i),\qquad
z_i^{R}=\mathrm{Slerp}(z_{S\rightarrow T},z_T,\alpha_i).
\end{equation}

Let $m=\lfloor N/2\rfloor$ be the midpoint index.
To reduce discontinuity at the frame handoff, we map the right midpoint back to the source frame with a weaker transport and fuse the two midpoint candidates, $\mathcal{T}_{1/2}$ denotes half-strength transport:
\begin{equation}
\tilde z_m^{R\rightarrow S}=\mathcal{T}_{1/2}(z_m^{R};\Phi^z_{T\rightarrow S},w_S^z),\qquad
z_m=\mathrm{Slerp}(z_m^{L},\tilde z_m^{R\rightarrow S},0.5).
\end{equation}
The initialized sequence is
\begin{equation}
z_i=
\begin{cases}
z_i^{L}, & i<m,\\
z_m, & i=m,\\
z_i^{R}, & i>m.
\end{cases}
\label{eq:aligned_init}
\end{equation}

For an attention layer $l$ at diffusion step $k$, the query $Q_{i,k}^l$ is computed from the current frame latent $z_{i}$ via the UNet hidden state at that layer. Hence, $Q_{i,k}^l$ is expressed in the same spatial coordinate frame as the current morphing latent. The key and value memories are constructed from two endpoint latents of the current phase. Given an endpoint pair $(z^{(1)}, z^{(2)})$, we form $(K_{i,k}^{(1)}, V_{i,k}^{(1)})$ and $(K_{i,k}^{(2)}, V_{i,k}^{(2)})$ from their corresponding UNet features at layer $l$ and step $k$. The attention output is then interpolated as:
\begin{equation}
\mathrm{Out}_{i,k}^l =
(1-\lambda_i)\,\mathrm{Attn}(Q_{i,k}^l,K_{i,k}^{(1)},V_{i,k}^{(1)})
+\lambda_i\,\mathrm{Attn}(Q_{i,k}^l,K_{i,k}^{(2)},V_{i,k}^{(2)}),
\label{eq:aligned_attn_mix}
\end{equation}
where $\lambda_i\in[0,1]$ is a monotone frame-wise schedule with $\lambda_0=0$ and $\lambda_{N-1}=1$.

To explicitly keep the query and memory in the same coordinate frame, inference is performed in two phases by instantiating $(z^{(1)}, z^{(2)})$ as follows:
For early frames ($i\le m$), $(z^{(1)}, z^{(2)}) = (z_S, z_{T\rightarrow S})$, aligning the memory with the source frame;
for late frames ($i>m$), $(z^{(1)}, z^{(2)}) = (z_{S\rightarrow T}, z_T)$, aligning the memory with the target frame.
This phase-wise construction ensures that $Q$ and both K/V branches are always defined in a consistent geometry.

\section{Experiments}
\label{sec:exp}

\subsection{Setup and Implementation Details}
\label{sec:exp_impl}

\textbf{Implementation Details.}
We implement AlignMorph on Stable Diffusion v2.1 using a single RTX 3090. Strictly following the FreeMorph protocol for fair comparison, we use a DDIM scheduler ($K=50$, edit strength $0.8$) for 40-step inversion and denoising. We generate $N=7$ frames at $768\times 768$ resolution with a CFG scale of 7.5, keeping all unmentioned configurations (e.g., seeds, noise initialization) identical to the baseline. For global semantic transport, DINOv2+FeatUp features are extracted and adaptively downsampled to $\le 16,384$ tokens for efficiency. We solve entropic optimal transport via log-domain Sinkhorn iterations ($\varepsilon=0.1$, 100 steps). Reliability masks use a cycle-consistency threshold $\tau=8$ pixels and a minimum confidence of 0.2. In multiband transport, low-frequency components are extracted via average pooling ($k=9$, stride 1, padding 4). The latent warping operates via bilinear grid sampling with border padding.

\begin{figure*}[t!]
    \centering
        \includegraphics[width=\textwidth]{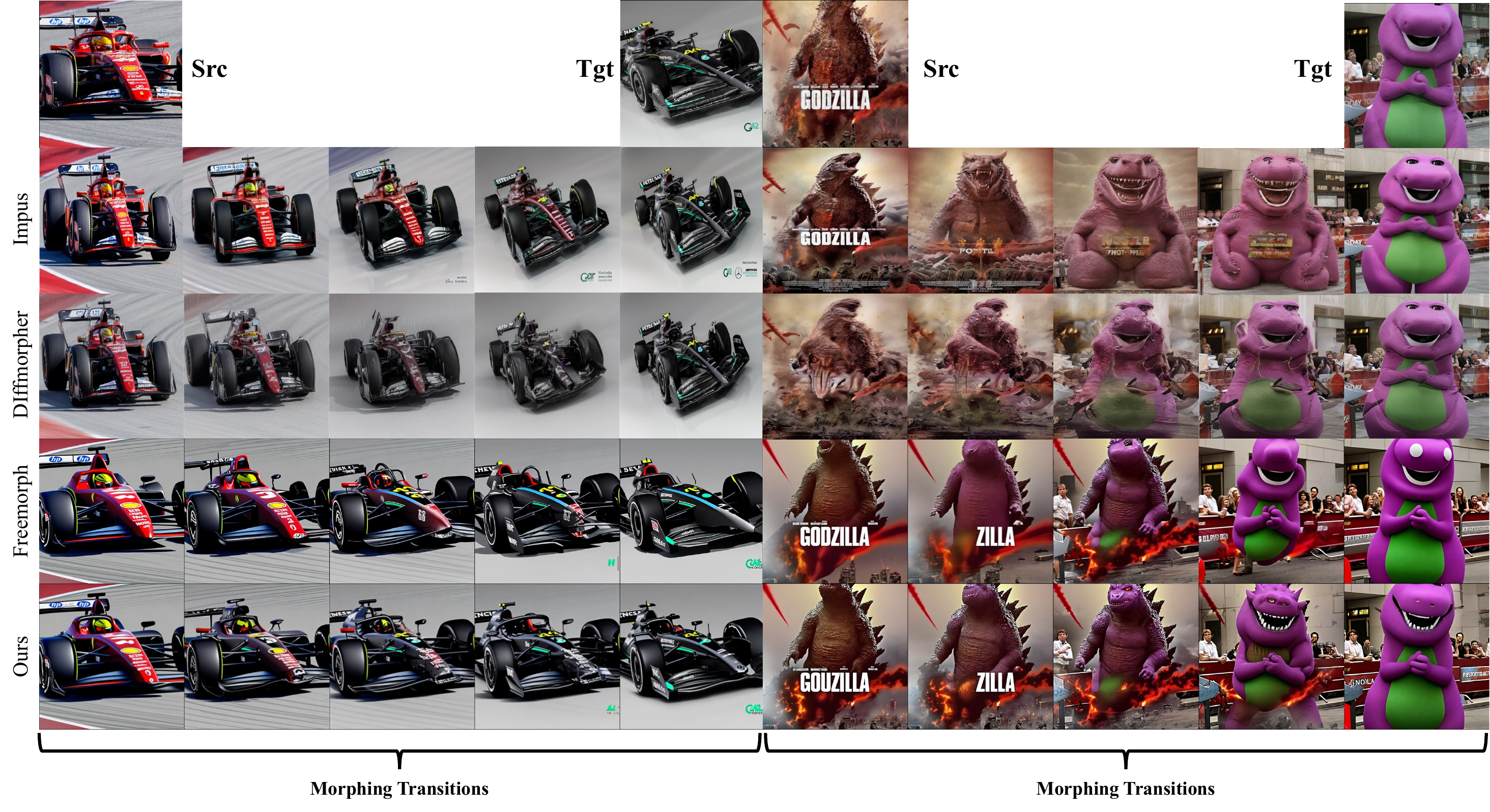}
\caption{
Qualitative comparison. Left (F1 car), Right (Godzilla).
}
    \label{fig:qual_compare}
\end{figure*}

\subsection{Quantitative Evaluations}
\label{sec:exp_quant}

Following prior work~\cite{yang2023impus,zhang2024diffmorpher,cao2025freemorph}, we evaluate fidelity and transition quality using three metrics:
(1) \textbf{FIDmean},
(2) \textbf{LPIPSsum} between adjacent frames, and
(3) \textbf{PPLsum} between adjacent frames.
Lower is better for all metrics.

Table~\ref{tab:quant} reports the results on MorphBench and Morph4Data.
AlignMorph achieves superior performance compared with both optimization-based and tuning-free baselines. In addition to quality metrics, we compare runtime efficiency.
AlignMorph requires $115$s per image pair on a single RTX 3090 GPU, including $25$s for offline correspondence and $90$s for diffusion sampling.
This runtime is comparable to other tuning-free methods such as FreeMorph ($90$s) under the same sampling schedule. In contrast, optimization-based approaches (e.g., IMPUS and DiffMorpher) require per-pair fine-tuning or latent optimization, typically taking tens of minutes per pair.
AlignMorph avoids any optimization and achieves approximately $20\times$–$50\times$ speedup over optimization-based methods.

\textbf{User Study.} To evaluate perceptual quality, we conducted a user study with 25 participants, ranging from university students to AI researchers. Each participant was presented with 50 randomly selected test cases and asked to choose the best result based on structural integrity and smoothness. As shown in Table~\ref{tab:user_study}, AlignMorph outperforms all baselines with a preference rate of \textbf{50.9\%}, confirming its superiority in generating natural and consistent transitions.

\begin{table}[t]
\centering
\small
\setlength{\tabcolsep}{4pt}
\caption{Quantitative comparison on MorphBench and Morph4Data. Results for the first three baselines are directly cited from~\cite{cao2025freemorph}. For rigorous benchmarking and to ensure a strictly fair comparison, we additionally report the reproduced results of FreeMorph using their official implementation. $\downarrow$ indicates lower is better.}
\resizebox{\textwidth}{!}{%
\begin{tabular}{l|ccc|ccc|ccc}
\hline
 & \multicolumn{3}{c|}{MorphBench} & \multicolumn{3}{c|}{Morph4Data} & \multicolumn{3}{c}{Overall} \\
Method & LPIPS$\downarrow$ & FID$\downarrow$ & PPL$\downarrow$
       & LPIPS$\downarrow$ & FID$\downarrow$ & PPL$\downarrow$
       & LPIPS$\downarrow$ & FID$\downarrow$ & PPL$\downarrow$ \\
\hline
IMPUS~\cite{yang2023impus}               & 130.52 & 152.43 & 3263.03 & 134.88 & 210.66 & 3199.90 & 265.40 & 174.76 & 6462.93 \\
DiffMorpher~\cite{zhang2024diffmorpher} & 90.57  & 157.18 & 2264.20 & 98.56  & 292.54 & 2394.05 & 189.13 & 209.10 & 4658.25 \\
FreeMorph~\cite{cao2025freemorph}       & 84.91  & 141.32 & 2122.80 & 80.30  & 201.09 & 2007.52 & 165.21 & 152.88 & 4130.32 \\
FreeMorph (Reproduced)                  & 85.01  & 139.79 & 2145.37 & 79.56  & 203.32 & 1993.42 & 164.57 & 154.45 & 4138.79 \\
\hline
\textbf{AlignMorph (ours)}              & \textbf{79.55} & \textbf{131.98} & \textbf{2010.63}
                                       & \textbf{74.23} & \textbf{188.44} & \textbf{1896.30}
                                       & \textbf{153.78} & \textbf{140.12} & \textbf{3896.93} \\
\hline
\end{tabular}%
}
\label{tab:quant}
\end{table}

\begin{figure}[t]
\centering
\includegraphics[width=\linewidth]{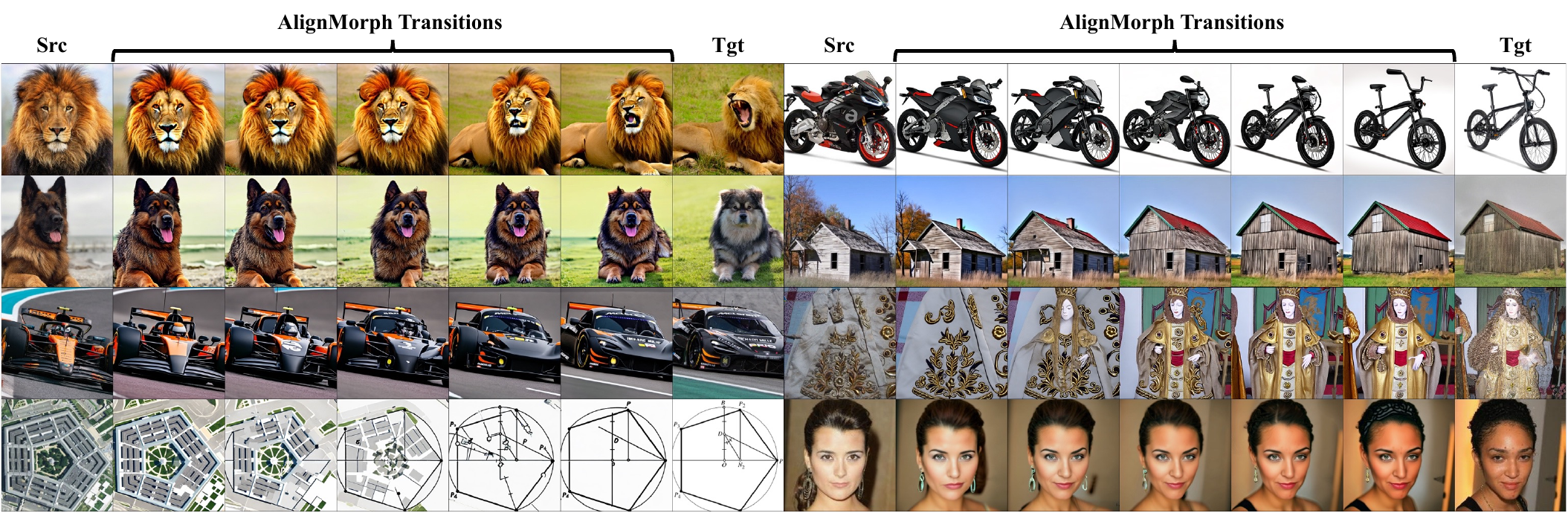}
\caption{More transitions generated by AlignMorph. Our method produces robust transitions across diverse image pairs under different scenarios.}
\label{fig:quali}
\end{figure}

\subsection{Qualitative Evaluations}
\label{sec:exp_qual}

\textbf{Qualitative Comparisons.}
As shown in \cref{fig:qual_compare}, we evaluate AlignMorph against baselines under large layout discrepancies.
Optimization-based methods (IMPUS, DiffMorpher) struggle with structural consistency, producing non-smooth transitions and hallucinating unrealistic artifacts entirely absent from the inputs.
While the tuning-free baseline FreeMorph mitigates these severe hallucinations, it still suffers from its implicit spatial alignment. Specifically, it exhibits temporally inconsistent color shifts and structural distortions in the F1 car (left). For the Godzilla morph (right), FreeMorph loses high-frequency details, resulting in blurry textures and intermediate frames with collapsed, meaningless semantics.
In contrast, by explicitly aligning geometry prior to denoising, AlignMorph successfully maintains crisp textures, coherent structural boundaries, and semantically meaningful intermediate states without any blending artifacts.

\textbf{Qualitative Results.}
We show additional morphing sequences generated by AlignMorph in \cref{fig:quali}.
Across diverse scenes and endpoint misalignment, AlignMorph produces coherent object transport and temporally stable synthesis.

\subsection{Ablation Study}
\label{sec:exp_analysis}

\begin{figure*}[t]
\centering
\begin{minipage}[t]{0.48\textwidth}
    \centering

    \captionof{table}{Ablation study. Lower is better.}
    \label{tab:ablation}
    \resizebox{\linewidth}{!}{
        \begin{tabular}{lccc}
            \hline
            Variant & LPIPS$\downarrow$ & FID$\downarrow$ & PPL$\downarrow$ \\
            \hline
            FreeMorph & 80.30 & 201.09 & 2007.52 \\
            Random OT & 132.14 & 297.60 & 2803.76 \\
            w/o Multiband & 88.16 & 231.97 & 2255.20 \\
            w/o Handoff & 91.70 & 257.91 & 2670.11 \\
            w/o KV warp & 76.12 & 193.76 & 1940.46 \\
            \hline
            AlignMorph & \textbf{74.23} & \textbf{188.44} & \textbf{1896.30} \\
            \hline
        \end{tabular}
    }

    \captionof{table}{User Preference Rate (\%).}
    \label{tab:user_study}
    \resizebox{\linewidth}{!}{
        \begin{tabular}{lcccc}
            \hline
            Method & DiffMorpher & FreeMorph & IMPUS & \textbf{Ours} \\
            \hline
            Preference & 8.2 & 26.4 & 14.5 & \textbf{50.9} \\
            \hline
        \end{tabular}
    }
\end{minipage}
\hfill
\begin{minipage}[t]{0.48\textwidth}
    \centering
    \vspace{0pt}
    \includegraphics[width=\textwidth]{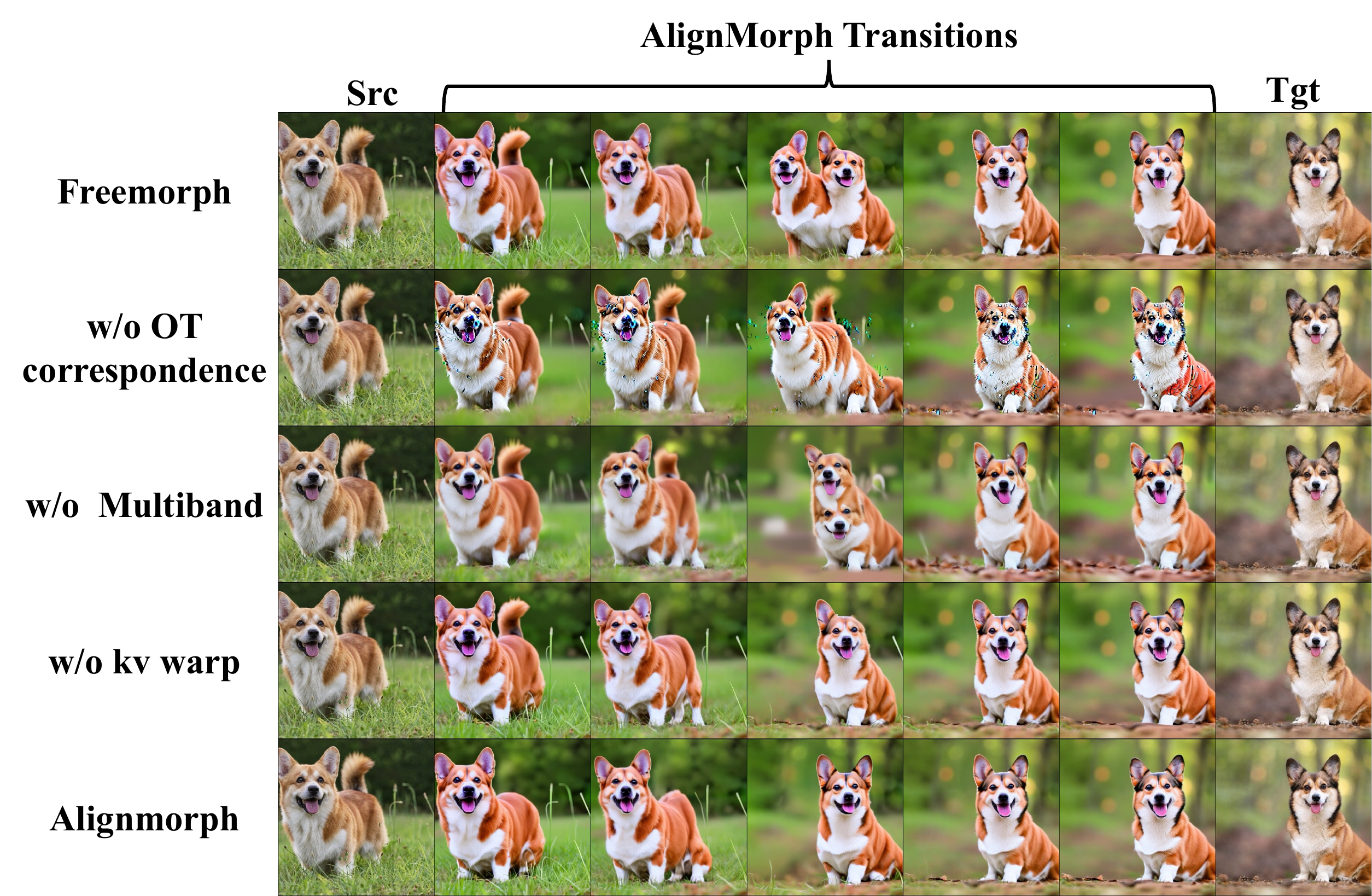} %
    \captionof{figure}{Qualitative comparison. AlignMorph preserves structural integrity better than baselines.}
    \label{fig:ablation}
\end{minipage}
\end{figure*}

This section analyzes the effects of AlignMorph components and connects them to the failure modes of tuning-free interpolation under large deformation. We conduct ablation experiments on Morph4Data to analyze the contribution of each component in AlignMorph.
Results are summarized in Table~\ref{tab:ablation}.

\textbf{w/o semantic correspondence.}
Replacing OT-based matching with a random mapping (20\% random displacement) degrades all metrics (Table~\ref{tab:ablation}). Visually (\cref{fig:ablation}), it introduces severe scattered noise and dark artifacts. This occurs because semantically unaware mapping blindly warps incompatible background features into the foreground, corrupting local latent statistics and proving the necessity of accurate semantic transport.

\textbf{w/o multiband latent transport.}
Removing the frequency-aware decoupling and directly warping the entire latent space results in a noticeable performance drop. Visually, also(\cref{fig:ablation}), textures become visibly blurred, and fine details in challenging regions (e.g., fur) are severely weakened. This validates the necessity of preserving high-frequency noise statistics. (We also verified that rigorous frequency decoupling via Fast Fourier Transform yields comparable generative quality, justifying our choice of spatial average pooling for its computational efficiency and simplicity.)

\textbf{w/o bi-phase coordinate handoff.}
In this variant, we remove the two-phase coordinate schedule and keep the entire morphing process in a single coordinate frame.
Specifically, the target representation is continuously warped into the source coordinate system, and no world-frame switch is performed at the morph midpoint.
As a result, the trajectory remains constrained within the source-aligned geometry and cannot fully transition into the target coordinate frame. The proposed bi-phase handoff resolves this issue by explicitly switching the world coordinate frame at the midpoint, allowing the trajectory to progressively adopt the target geometry while maintaining consistent transitions.

Empirically, we observe that warping the \emph{UNet input latent} (from which the query $Q$ is derived) serves as the primary driver for geometric alignment. Even when endpoint memories $(K,V)$ remain unwarped (\textbf{w/o KV warp}), our bi-phase handoff framework still generates structurally coherent transitions. This behavior aligns with the fundamental nature of attention as a coordinate-sensitive retrieval operator: $Q$ dictates \emph{where} to retrieve, while $(K,V)$ dictate \emph{what} is retrieved. Under large deformations, the dominant failure mode is a \emph{query-side spatial mismatch}---queries are spatialized on an unaligned intermediate grid, whereas endpoint memories reside in a fixed endpoint frame. Warping the input latent explicitly aligns the \emph{query-side coordinate frame} prior to $QK^\top$ matching, fundamentally resolving severe structural entanglement. Subsequently, aligning $(K,V)$ ensures the \emph{content consistency} of the retrieved features against the transported geometry, effectively eliminating residual texture drift and minor ghosting. Thus, while $Q$ alignment establishes the macroscopic geometric structure, $K\!V$ alignment provides a vital complementary refinement that maximizes fine-grained texture fidelity and overall visual quality.

\section{Limitations and Future Work}

\begin{figure}[t]
\centering
\includegraphics[width=\linewidth]{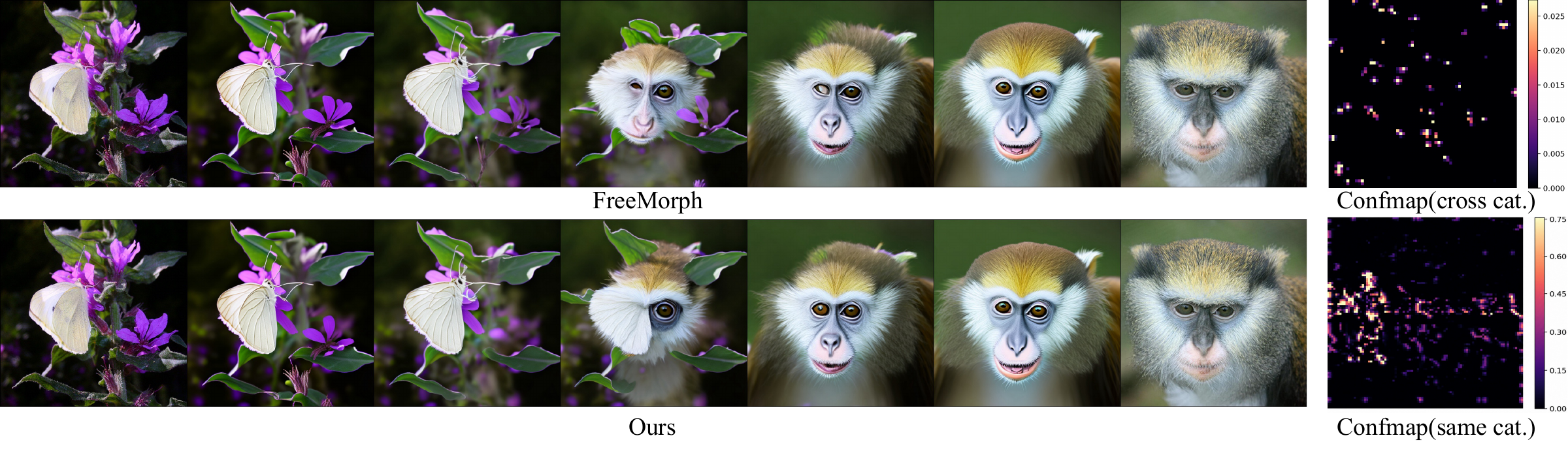}
\vspace{-8pt}
\caption{Failure analysis on extreme cross-category morphing. Left: comparison with FreeMorph on a flower-to-monkey pair. Right: confidence maps for cross-category and same-category pairs.}
\label{fig:failurecase}
\end{figure}

Current morphing benchmarks primarily focus on image pairs with partial semantic or structural overlap, where meaningful correspondence and geometric alignment are well-defined. As shown in Fig.~\ref{fig:failurecase}, extreme cross-category pairs instead expose a different regime: the semantic correspondence itself becomes invalid. For example, in flower-to-monkey morphing, there is no canonical mapping between petals, leaves, and facial parts. This is reflected by the confidence map, where the cross-category case produces consistently low responses. In contrast, when the two endpoints belong to the same category but undergo large spatial displacement, the confidence map remains much stronger, indicating that valid part-level correspondence can still be established. Accordingly, our reliability gate suppresses unreliable transport in the former case rather than enforcing arbitrary OT warps. AlignMorph therefore degrades toward an interpolation-dominant trajectory, which avoids additional wrong-warp artifacts but also reduces the advantage of explicit alignment over interpolation-based transitions.

This limitation is different from large deformation with valid semantic correspondence. For same-category pairs, the confidence map remains much denser, indicating that reliable part-level correspondence can still be established despite large layout displacement. Therefore, the main boundary of AlignMorph is not deformation magnitude, but the existence of meaningful semantic correspondence. Extending morphing beyond shared structural similarity requires redefining the alignment objective and perceptual criteria for open-category generative blending, which we leave for future work.

\section{Conclusion}
In this paper, we presented \textbf{AlignMorph}, a novel tuning-free diffusion morphing framework that tackles the severe structural entanglement prevalent under large layout discrepancies. Guided by the \emph{transport-then-denoise} paradigm, we explicitly decouple geometric alignment from generative refinement. Specifically, our framework first employs Global Semantic Transport to establish dense semantic correspondences via optimal transport, which are then projected into the latent space through a frequency-aware multiband transport. This crucial step guarantees reliable macro-geometric alignment while strictly preserving the high-frequency noise statistics inherent to diffusion priors. Subsequently, Coordinate-Aligned Generation synthesizes the intermediate transitions via a symmetric bi-phase attention handoff, ensuring strict spatial consistency throughout the denoising process. Extensive experiments demonstrate that AlignMorph effectively eliminates ghosting artifacts and achieves superior performance without the need for expensive per-pair optimization. We hope this work highlights the necessity of explicit geometric alignment in diffusion-based generation and inspires future research into more robust generative transport mechanisms.

\section*{Acknowledgements}

This work was supported by the National Key Research and Development Program for Young Scientists of China (No. 2024YFB3310100), the Natural Science Foundation of Hubei Province (No. 2025AFB592), and the Natural Science Foundation of Wuhan (No. 2025040601020216).

\bibliographystyle{splncs04}
\bibliography{main}
\end{document}